\documentclass[dvipsnames,authorversion,nonacm,format=sigconf]{acmart}
\AtBeginDocument{%
	\providecommand\BibTeX{{%
			\normalfont B\kern-0.5em{\scshape i\kern-0.25em b}\kern-0.8em\TeX}}}

\usepackage{subcaption}
\usepackage{algorithm2e}
\usepackage{graphicx}
\usepackage{textcomp}
\usepackage{xcolor}
\usepackage{url}
\def\BibTeX{{\rm B\kern-.05em{\sc i\kern-.025em b}\kern-.08em
    T\kern-.1667em\lower.7ex\hbox{E}\kern-.125emX}}

\usepackage{algpseudocode}% http://ctan.org/pkg/algorithmicx

\usepackage[acronym]{glossaries}
\usepackage{xspace}
\graphicspath{{figures/}}

\newcommand{\MP}[1]{\mbox{{\bf MUX#1}}}

\newcommand{\MAJ}[1]{\mbox{{\bf MAJ#1}}}

\newcommand{\W}[1]{\mbox{\bf Woods#1}}
\newcommand{\MAZE}[1]{\mbox{\bf Maze#1}}

\newcommand{\cl}{\mbox{\emph{cl}}}

\newcommand{\CANP}{$\text{\it CAN}(P)$}
\newcommand{\MANP}{$\text{\it MAN}(P)$}
\newcommand{\CAN}{\text{\it CAN}}
\newcommand{\MAN}{\text{\it MAN}}

\begin{document}

\title{An Approach to Analyze Niche Evolution in XCS Models}

\author{Pier Luca Lanzi}
%\authornote{Both authors contributed equally to this research.}
\email{pierluca.lanzi@polimi.it}
\orcid{0000-0002-1933-7717}
\affiliation{%
	\institution{Politecnico di Milano}
%	\streetaddress{P.O. Box 1212}
	\city{Milano}
%	\state{Ohio}
	\country{Italy}
%	\postcode{43017-6221}
}

%\author{Daniele Loiacono}
%%\authornote{Both authors contributed equally to this research.}
%\email{daniele.loiacono@polimi.it}
%\orcid{0000-0002-5355-0634}
%\affiliation{%
%	\institution{Politecnico di Milano}
%	%	\streetaddress{P.O. Box 1212}
%	\city{Milano}
%	%	\state{Ohio}
%	\country{Italy}
%	%	\postcode{43017-6221}
%}

%\author{Ben Trovato}
%\authornote{Both authors contributed equally to this research.}
%\email{trovato@corporation.com}
%\orcid{1234-5678-9012}
%\author{G.K.M. Tobin}
%\authornotemark[1]
%\email{webmaster@marysville-ohio.com}
%\affiliation{%
%	\institution{Institute for Clarity in Documentation}
%	\streetaddress{P.O. Box 1212}
%	\city{Dublin}
%	\state{Ohio}
%	\country{USA}
%	\postcode{43017-6221}
%}

%\author{Ben Trovato}
%\authornote{Both authors contributed equally to this research.}
%\email{trovato@corporation.com}
%\orcid{1234-5678-9012}
%\author{G.K.M. Tobin}
%\authornotemark[1]
%\email{webmaster@marysville-ohio.com}
%\affiliation{%
%	\institution{Institute for Clarity in Documentation}
%	\streetaddress{P.O. Box 1212}
%	\city{Dublin}
%	\state{Ohio}
%	\country{USA}
%	\postcode{43017-6221}
%}
%
%\author{
%	\IEEEauthorblockN{Elio Sasso}
%	\IEEEauthorblockA{\textit{LaborA Model Laboratory}}\\
%	\textit{Politecnico di Milano}\\
%	Milano, Italy\\
%	elio.sasso@polimi.it
%\and
%	\IEEEauthorblockN{Daniele Loiacono}
%	\IEEEauthorblockA{\textit{DEIB}} \\
%	\textit{Politecnico di Milano}\\
%	Milano, Italy\\
%	daniele.loiacono@polimi.it
% \and
%	\IEEEauthorblockN{Pier Luca Lanzi}
%	\IEEEauthorblockA{\textit{DEIB}} \\
%	\textit{Politecnico di Milano}\\
%	Milano, Italy\\
%	pierluca.lanzi@polimi.it
%}

%%% REVISED 2023-01-09
\begin{abstract}
We present an approach to identify and track the evolution of niches in XCS that can be applied to any XCS model and any problem. It exploits the underlying principles of the evolutionary component of XCS, and therefore, it is independent of the representation used. It also employs information already available in XCS and thus requires minimal modifications to an existing XCS implementation. We present experiments on binary single-step and multi-step problems involving non-overlapping and highly overlapping solutions. We show that our approach can identify and evaluate the number of niches in the population; it also show that it can be used to identify the composition of active niches to as to track their evolution over time, allowing for a more in-depth analysis of XCS behavior.
\end{abstract}

%%%
%%% The code below is generated by the tool at http://dl.acm.org/ccs.cfm.
%%% Please copy and paste the code instead of the example below.
%%%
%\begin{CCSXML}
%<ccs2012>
%   <concept>
%       <concept_id>10010147.10010257.10010293.10011809.10011812</concept_id>
%       <concept_desc>Computing methodologies~Genetic algorithms</concept_desc>
%       <concept_significance>500</concept_significance>
%       </concept>
%   <concept>
%       <concept_id>10003752.10003753.10003759</concept_id>
%       <concept_desc>Theory of computation~Interactive computation</concept_desc>
%       <concept_significance>500</concept_significance>
%       </concept>
%   <concept>
%       <concept_id>10003120.10003123.10011760</concept_id>
%       <concept_desc>Human-centered computing~Systems and tools for interaction design</concept_desc>
%       <concept_significance>500</concept_significance>
%       </concept>
% </ccs2012>
%\end{CCSXML}

%\ccsdesc[500]{Computing methodologies~Genetic algorithms}
%\ccsdesc[500]{Theory of computation~Interactive computation}
%\ccsdesc[500]{Human-centered computing~Systems and tools for interaction design}

%%
%% Keywords. The author(s) should pick words that accurately describe
%% the work being presented. Separate the keywords with commas.

\maketitle

\section{Introduction}
%Learning classifier systems are methods of genetics-based machine learning that combine a rule-based representation, reinforcement learning, and evolutionary computation to solve classification, regression, and reinforcement learning problems. They maintain a population of condition-action-prediction rules (the classifiers) that represent the solution to a target problem. Reinforcement learning is applied to update the classifier prediction based on how valuable the action is for solving the problem (the higher the prediction, the more valuable the action). Evolutionary computation selects, reproduces, recombines, and mutates the most promising classifiers. Early learning classifier systems were focused on classifier prediction (or strength), used by both the reinforcement (credit assignment) and the evolutionary components. 

XCS \cite{wilson:1995} has become the state-of-the-art in genetics-based machine learning as most, if not all, learning classifier models introduced nowadays are based on its three principles. In XCS, the classifier prediction measures how valuable the classifier action is in terms of problem solution;
the classifier fitness evaluates how accurate the prediction is. The genetic algorithm selects, recombines, and mutates accurate classifiers in niches formed by classifiers that advocate the same action in similar situations. More general classifiers will appear in more niches and, if they are accurate, they will reproduce more. As a result, XCS evolves maximally general, maximally accurate populations representing a complete mapping of the optimal solution in which classifiers represent local solutions to meaningful problem subspaces \cite{wilson:1998}. 

XCS has two components working together but also sort of separately. On the one hand, classifiers’ predictions are continuously updated based on the value of their actions in terms of solution. On the other hand, fitness values are updated to estimate the accuracy of such varying prediction values, while the genetic algorithm uses such partial information to evolve better rules. Thus, XCS is learning a prediction landscape to estimate the value of actions in every possible situation; meanwhile, it is also building a fitness landscape to guide the evolution of better rules. This is similar to reinforcement learning with function approximators, where systems learn an optimal action-value function while computing the intermediate solutions' best approximation.

\vfill\null
\medskip\noindent\textbf{Related Work.} 
The analysis of the evolution of classifier populations is challenging since the two components are interconnected, and their workings also depend on the representation of classifier conditions. Thus, learning is usually analyzed using proxies such as the reward the system receives, the number of actions needed to reach a solution, and the number of classifiers in the population. 
%
%In-depth analyses are performed using the traditional ternary representation and the knowledge of the optimal solution \cite{wilson:1998}.
Mathematical models have been developed to explain specific aspects of the learning and evolution in XCS \cite{10.1145/3319619.3326848}, for instance the population sizing to guarantee that all the required evolutionary niches are represented and maintained in the population \cite{lanzi:2001:GECCOhow,DBLP:journals/ec/ButzGT03,lanzi:2004:tec}; the time to convergence \cite{DBLP:conf/gecco/ButzGL04}; the balance between occurrence-based activation and support-based deletion \cite{DBLP:journals/gpem/ButzGLS07}; the frequency of genetic algorithm activations and the error thresholds needed to guarantee that accurate solution can be evolved \cite{DBLP:journals/tec/NakataB21,DBLP:journals/access/SugawaraN22}. However, all these models are usually validated using the ternary representation and the knowledge of how the optimal solution is represented using such basic representation. When studying evolution in XCS with more advanced representation (e.g., real-coded intervals \cite{DBLP:conf/iwcls/Wilson99a,DBLP:journals/ec/StoneB03}, code fragments \cite{6595603}, genetic programming trees \cite{lanzi:1999:GECCO:gp,DBLP:conf/cec/Lanzi07}), the analyses are based on population-level statistics and the application of condition simplification \cite{DBLP:conf/cec/Lanzi07}, \cite{DBLP:journals/telo/LiuBX21}, and visualization \cite{DBLP:journals/nc/LiuBX22} algorithms. Recently, \citet{lanzi:2025:optimal} introduced an approach to compute the optimal solution for multi-step problems based on the theory of generalization in XCS \cite{lanzi:2001:GECCOhow,DBLP:journals/ec/ButzGT03,lanzi:2004:tec,DBLP:journals/gpem/ButzGLS07} and logic minimization that for the first time provided the optimal solution to well-known binary multi-step problems \cite{wilson:1995,wilson:1998,lanzi:1999:analysis,lanzi:2004:tec}.

\medskip\noindent\textbf{Contribution.} 
In this paper, we present an approach to identifying and tracking the currently active niches in XCS evolving populations 
	that exploits its occurrence-based activation and reproduction mechanism. 
Our approach is independent of the representation and does not require knowledge about the problem solution used in other analyses \cite{wilson:1998,lanzi:2004:tec,DBLP:journals/gpem/ButzGLS07}. It also employs information readily available in XCS, 
thus requiring minimal modifications to any XCS implementation. 
We present experimental results on binary single-step and multi-step problems involving non-overlapping and highly overlapping optimal solutions. We show that our approach can provide a robust estimate of the number of active niches in the evolving populations. We also show that our approach can be applied to identify and trace the evolution of niches over time.

\section{The XCS Classifier System}
\label{sec:xcs}
We provide a brief overview of XCS focused on the relevant elements for the discussion in this paper. We refer the reader to \cite{DBLP:journals/soco/ButzW02} for a detailed description.

\subsection{The Performance Component}
XCS maintains a population of rules (classifiers), which represents the current solution to a problem. Classifiers have a condition, an action, and two main parameters: (i) the prediction $p$, estimating the payoff the system expects when the classifier is used;
the fitness $f$, estimating the accuracy of the prediction $p$. 

At time $t$, XCS builds the match set [M] with the classifiers in the population whose condition matches the current sensory input $s_t$; if [M] does not contain classifiers for all the feasible actions, it applies covering and generates random classifiers that match the current input and have the missing actions. For each possible action $a$, XCS computes 
the system prediction $P(s_t, a)$ as the fitness weighted average of the predictions of classifiers in [M] that advocate action $a$ \cite{wilson:1995}. $P(s_t, a)$ is an estimation of the payoff that the XCS expects if action $a$ is performed in $s_t$. Next, XCS selects an action to perform and the classifiers in [M] with the selected action are added to the action set [A]. The chosen action $a_t$ is performed and XCS receives a reward $r_{t+1}$ and 
a new input $s_{t+1}$. The reward is used to update the parameters of the classifiers that are deemed accountable for it. Every time the parameters of a classifier are updated, its experience parameter (\textit{exp} in \cite{DBLP:journals/soco/ButzW02}) is increased to keep track of how reliable the parameters are.

\subsection{The Evolutionary (Discovery) Component}
\label{ssec:discovery}
On a regular basis, the genetic algorithm is applied to the action set [A], which represents the evolutionary niche of classifiers that, in a particular context, invoke the same action \cite{wilson:1998,lanzi:2004:tec}. The (steady-state) genetic algorithm selects two classifiers, copies them, performs crossover on the copies with probability $\chi$, and mutates them with probability $\mu$. The offspring classifiers are inserted into the population and two classifiers are deleted to keep the population size constant.

The activation of the genetic algorithm is controlled by the parameter $\theta_{ga}$ that defines how much time  should pass (on average) between activations in the same action set  [A]. For this purpose, classifiers in XCS have a timestamp parameter $ts$ that is updated to the current time $t$ every time the genetic algorithm is applied in the action set [A]. XCS computes the average timestamp of the classifiers in  [A]. If the difference between the current time $t$ and the average $ts$ is greater than $\theta_{ga}$, then enough time has passed since the last activation, so the genetic algorithm is applied to the classifiers in [A]. 

\section{Active Evolutionary Niches}
\label{sec:identifying}
We describe our approach to identify active evolutionary niches in XCS using a simple binary problem. 

\subsection{A Simple Binary Problem}
Consider the following classification problem with a 5-bit input and a binary action. 
XCS receives a reward of 1000 when applying the action corresponding to the first input bit, zero otherwise;  
for action 0, XCS will receive a 0 reward for input \texttt{10000} and a 1000 reward for input \texttt{01010};
for the same inputs, action 1 will receive a reward of 1000 and 0 respectively. 
Table \ref{tab:example} shows the optimal (maximally general, maximally accurate) solution that XCS can evolve for this problem where $O(d,a)$ identifies the optimal classifier with action $a$ that matches the first digit $d$. To evolve this solution, XCS must maintain four evolutionary niches in the population at all time, each one identified by the optimal classifiers in Table \ref{tab:example} \cite{lanzi:2004:tec}.  

\begin{table}[t]
	\begin{tabular}{|c|c|c|}\hline
		\texttt{Classifier} & \texttt{Condition:Action} & \texttt{Prediction}\\ \hline\hline
		$O(0,0)$ & \texttt{0\#\#\#\#:0} & 	1000\\
		$O(0,1)$ & \texttt{0\#\#\#\#:1} & 0 \\
		$O(1,0)$& \texttt{1\#\#\#\#:0} & 0 \\
		$O(1,1)$ & \texttt{1\#\#\#\#:1} & 1000\\\hline
	\end{tabular}
	\caption{Optimal solution for the example problem.}
	\label{tab:example}
\end{table}

\subsection{Modifications to XCS Models}
\label{ssec:modifications}
Our approach requires two simple modifications that can be applied to any existing XCS model. First, we extend the classifier structure with an \textit{action set time stamp} (\textit{ats}), storing the last time the classifier was added to an action set, and a list LIFO $L$ of previous classifier's \textit{ats} values. When a classifier is created, through covering or by the genetic algorithm, \textit{ats} is set to zero and $L$ is empty. Next, we modify XCS code so that when a classifier is added in the action set (function \textit{GENERATE ACTION SET} in \cite{DBLP:journals/soco/ButzW02}), the classifier's \textit{ats} is updated to the current time and its value is pushed in the first position of $L$.

\subsection{Number of Currently Active Niches}
Suppose that XCS has already reached the optimal solution and the population contains only several copies of  the four classifiers in Table~\ref{tab:example}. Let us denote with $t(d,a)$ the last time the input state had first digit $d$ and action $a$ was applied. Every time a classifier is added into the action set (Section~\ref{ssec:modifications}), its \textit{ats} is updated to the current time; accordingly, the time stamp is a monotonically increasing parameter that indicates the last time a classifier was applied. If we compute the set of \textit{ats}  values $T$ for all the classifiers in the population, for any population size, we simply obtain
\[T = \{t(0,0), t(0,1), t(1,0), t(1,1)\}\] 
Let us now assume that XCS is still learning so that the population contains both more specific but accurate classifiers and overgeneral ones matching more niches. At a given time, the population can be partitioned into the six classifier sets $C(d,a)$ containing the classifiers whose condition starts with a symbol $d$ in $\{0,1,\#\}$ and have action $a$ in $\{0,1\}$. These classifier sets correspond to the niches that XCS maintains while learning to solve the task. The time stamps of all the \textit{accurate} classifiers in $C(0,0)$, $C(0,1)$, $C(1,0)$ and $C(1,1)$ will be $t(0,0)$, $t(0,1)$, $t(1,0)$, $t(1,1)$ respectively. The overgeneral classifiers in $C(\#,0)$ will be applied both when the input state starts with 0 or 1. Accordingly, their time stamp will be either $t(0,0)$ or $t(1,0)$ depending on what was the first digit of the latest input for which action 0 was applied, that is, 
\[
t(\#,0) = \max \{t(0,0), t(1,0)\}
\]
\noindent
similarly, $t(\#,1)$ will be $\max \{t(0,1), t(1,1)\}$. Thus, in a population the number of distinct \textit{ats} values estimates the number of active niches in the population. 

We can identify the set of \textit{Currently Active Niches} (\textit{CAN}) in the population $P$ using their corresponding \textit{ats} values, that is, 
\begin{equation}
	\text{\it CAN}(P) = \{\cl.ats\ |\ \cl \in P\} \label{eq:an}
\end{equation}
and compute the number of niches in the population as $|\text{\it CAN}(P)|$.
%\begin{equation}
%	\text{\it NoCAN}(P) = |\text{\it CAN}(P)| \label{eq:nn}
%\end{equation}
Note that, we distinguish between active and inactive niches since there might be freshly created classifiers that did not appear in an action set yet and therefore they cannot be assigned to a niche since they have not been applied yet. These classifiers should not be considered 
%when computing $\text{ActiveNiches}(P)$ 
and can be easily identified from their experience \textit{exp} (Section~\ref{ssec:discovery}) that counts the number of updates a classifier received and therefore it is zero for inactive classifiers \cite{wilson:1995,DBLP:journals/soco/ButzW02}

\subsection{Composition of Active Niches}
\label{ssec:composition}
The number of \textit{ats} values in the population estimates the number of active evolutionary niches, but we need more information to compute the composition of active niches. In the previous example, the time stamps in $C(\#,0)$ will be either $t(0,0)$ or $t(1,0)$ depending on the last input XCS received. If we look only at \textit{ats} values, we cannot uncover that the classifiers in $C(\#,0)$ belong to two evolutionary niches, $C(0,0)$ and $C(1,0)$. To derive this information, we must keep track of $C(\#,0)$ classifiers’ previous \textit{ats} values.

Suppose the first digit of the last input was 0; the time stamps in $C(\#,0)$ would be $t(0,0)$ with $t(0,0)>t(1,0)$ since the last input starting with 1 was received before the current input beginning with 0. If we keep list of previous time stamps assigned to each classifier in $C(\#,0)$, then $t(1,0)$ would be in such a list since at time $t(1,0)$, when XCS received an input starting with 1 and performed action 0, classifiers in $C(\#,0)$ belonged to the action set.

We can compute the composition of active evolutionary niches in XCS by adding a list of time stamps $L$ to each classifier. When the timestamps of the classifiers in the action set are updated, the new timestamp is added in the first position of each classifier list. For each active niche identified by a time stamp value \textit{ats} ($ats\in \text{ActiveNiches}(P)$), we can compute the classifiers in the population that belongs to it $\text{Niche}(ats,P)$ as, 
\[ \text{Niche}(ats, P) = \{\cl\ |\ \cl \in P \wedge ats \in \cl.L\}\]
That is the set of classifiers that have \textit{ats} in their timestamp list using the notation of \cite{DBLP:journals/soco/ButzW02}.

\subsection{The Case of Overlapping Solutions}
\label{ssec:overlapping}
\CANP\ provides a snapshots of all niches that have been recently visited. In problems with solutions involving non-overlapping classifiers like the Boolean multiplexer \MP{} (Section~\ref{sec:boolean_functions}), such snapshot is likely to identify all the existing niches. However, in problems with solutions mainly consisting of overlapping classifiers, like the majority-on functions \MAJ{} (Section~\ref{sec:boolean_functions}), such snapshot might underestimate the number of niches in the population.

\begin{figure}[t]
	\subfloat[]{
		\centering
		\includegraphics[width=0.8\columnwidth]{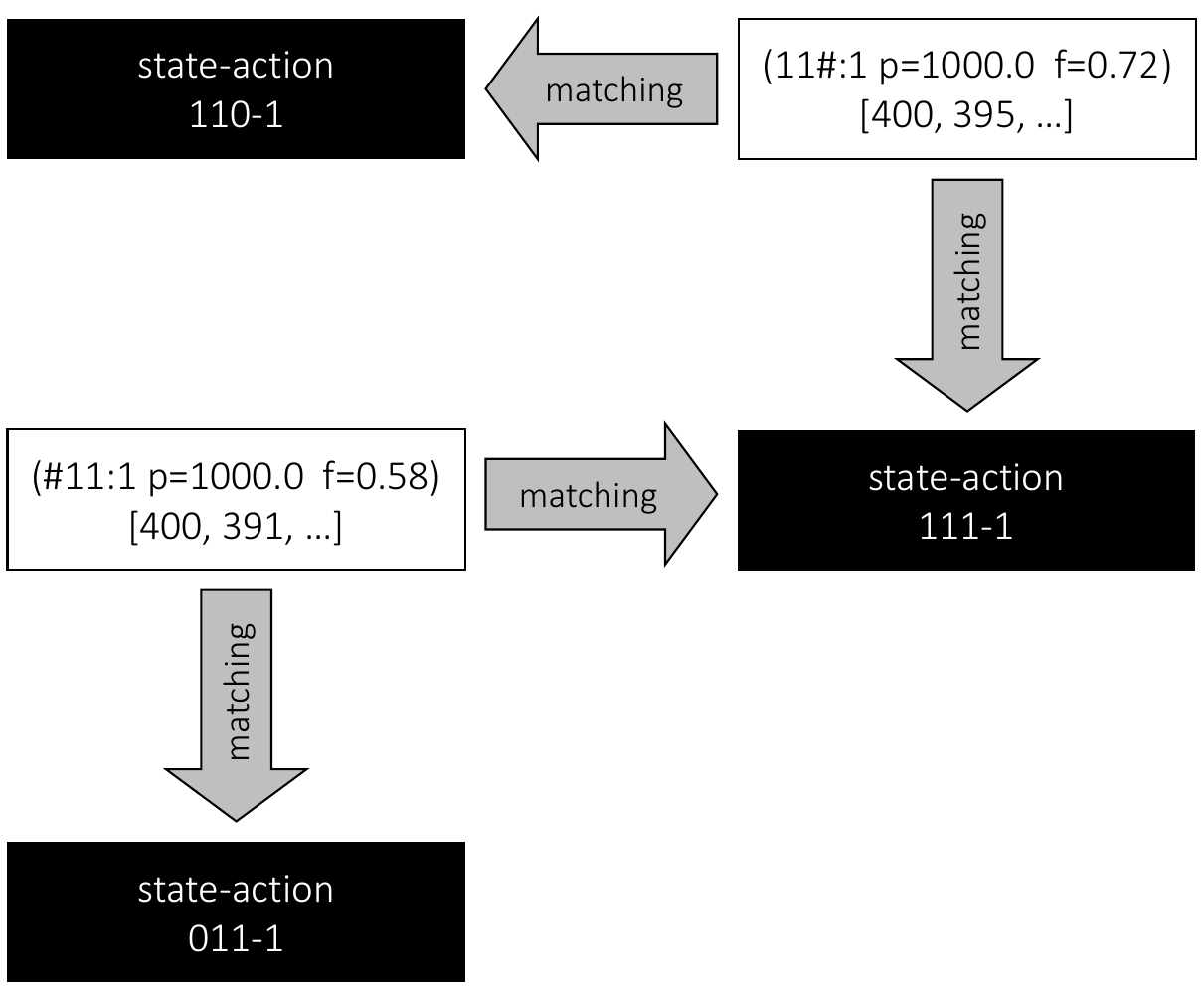}
		\label{fig:overlapping_niches_1}
	}

	\subfloat[]{
		\centering
		\includegraphics[width=0.8\columnwidth]{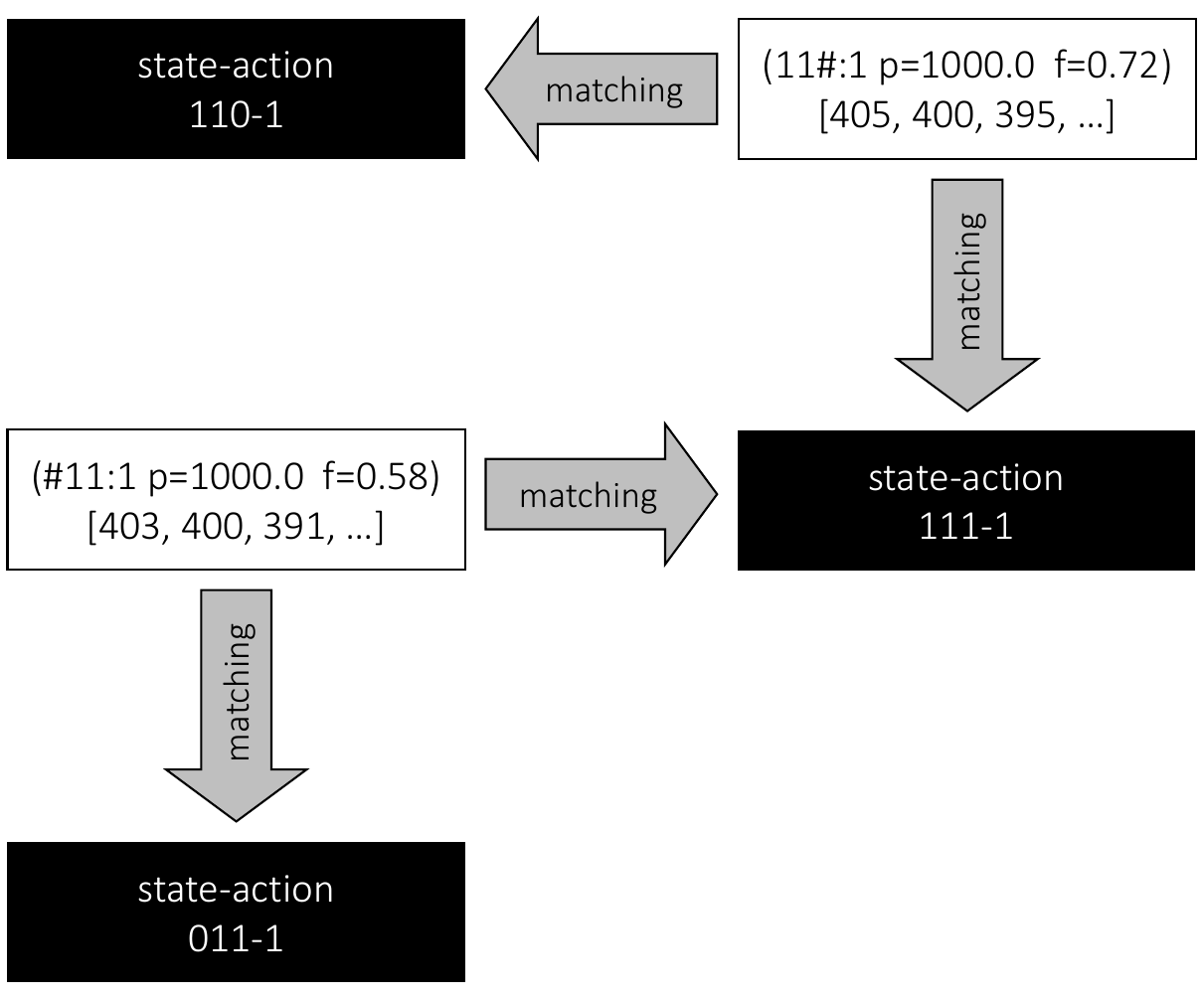}
		\label{fig:overlapping_niches_2}
	}
	
	\caption{Example of overlapping classifiers from an optimal solution for \MAJ{3}; the black boxes represent input-action pairs; the white boxes show classifiers using their condition:action pairs, prediction $p$, fitness $f$, and the content of the list $L$ of $ats$ values starting from the last one added.}
	\label{fig:overlapping_example}
%	\centerline{\includegraphics[0.8\columnwidth]{./figures/OverlappingNichesExamples-Fig1}}
\end{figure}

Figure~\ref{fig:overlapping_example} shows two overlapping classifiers evolved for the majority-on function of size 3 (\MAJ{3}) that are part of the optimal (accurate maximally general) population XCS evolved. The black boxes (labeled \mbox{110-1}, \mbox{111-1}, and \mbox{011-1}) identify three niches using input-action pairs; the white boxes represent the two overlapping classifiers using their \textit{condition:action} pairs (\mbox{11\#:1} and \mbox{\#11:1}), their prediction $p$, fitness $f$, and the content of the list $L$ of $ats$ values starting from the last $ats$ added (Section~\ref{ssec:modifications}). Figure~\ref{fig:overlapping_example}a shows the case in which both classifiers have been activated in \mbox{111-1} and thus their $ats$ values has been updated with the same niche identifier (400) which will has also been pushed into $L$. Accordingly, when considering the current number of niches in the population using only the classifiers' $ats$ values (the first element in $L$), \CANP\ will return only one niche (with $ats$ 400 in Figure~\ref{fig:overlapping_example}a) although the same classifiers are involved in three niches not one. 
Suppose now that in the next iterations (Figure~\ref{fig:overlapping_example}b), XCS first visits niche \mbox{011-1} and then \mbox{110-1} activating the two classifiers separately. In this case, their $ats$ values will be updated with different niche identifiers and \CANP\ will show two separate niches (with id 403 and 405 in Figure~\ref{fig:overlapping_example}b). Therefore, in general, we cannot rely only on the most recent $ats$ values but we need to consider also the previous niches in which classifiers have been activated, recorded in the classifiers' lists $L$.

In the example in Figure~\ref{fig:overlapping_example}b, we could identify all the niches in which the two classifiers are involved by comparing their $ats$ lists. Two classifiers that have the same niche identifier (400) in $L$ belongs to the same niches; however, if they also have mismatching identifiers (405 and 403) they also belong to these other niches. Thus, the total number of niches can be computed as the sum of number of common and mismatching niches. This procedure would return the exact number of niches in the population but unfortunately its complexity is combinatorial and therefore computationally infeasible in reasonable scenarios. 
Accordingly, we propose to \textit{estimate} the number of niches in the population as the average of the number of niches recorded in all the lists $L$ of the classifiers in the population. We define, 
\begin{equation}
	\text{\it CAN}_t(P) = \{\cl.L[t]\ |\ \cl \in P\} \label{eq:an_i}
\end{equation}
that is the set of all the $ats$ values stored in position $t$ and represents the snapshot of the active niches $t$ time steps ago. We compute the mean number of active niches and as, 
\begin{equation}
	\text{\it MAN}(P) = \frac{1}{|L|} \sum_{t=0}^{|L|-1} |\text{\it CAN}_t(P)|. \label{eq:man}
%	\text{\it CAN}$_t$(P) = \{\cl.L[t]\ |\ \cl \in P\} \label{eq:man}
\end{equation}
Note that \MANP\ provides an evaluation of the average number of the niches that are active in the recently recorded populations and therefore it might underestimate the actual number of niches. However, as we will see later in Section~\ref{sec:experiments_binary}, its value combined to the corresponding standard deviation provides a reasonable evaluation of the number of niches in the population at a limited computational cost.

\subsection{Discussion}
%Our approach requires to modify an XCS implementation (i) by adding the action set time stamp \textit{ats} and the list of time stamps $L$ to classifiers' structure; and (ii) by adding two lines of code to set \textit{ats} values and update time stamp list $L$ when classifiers are added to the action set. 
XCS classifiers already have the time stamp $ts$ for triggering the genetic algorithm activation (Section~\ref{sec:xcs}). A preliminary analysis we performed shows a strong correlation between $ts$ and $ats$ parameters. Thus, in principle, it might be possible to modify XCS to use only one timestamp for the genetic algorithm activation and the tracking of active niches. However, we preferred to introduce a separate parameter so that its addition to an existing implementation does not interfere with the current code.

An evolutionary niche needs at least one representative classifier. 
Accordingly, the time stamp list $L$ can contain at most as many values as the number of classifiers
	in the population $N$ \cite{wilson:1995,DBLP:journals/soco/ButzW02}.
However, \cite{lanzi:2004:tec} showed that XCS needs to sustain evolutionary niches with more classifiers to keep each niche alive. 
We performed a series of experiments with XCS using different sizes of $L$; 
	we determined that $L$ maximum size can be set to 10\% of $N$ without decreasing the tracking capability of our approach.
The experiments discussed in this paper use this setting.
%
%Butz et al. [1] proposed that LCSs are designed to evolve rulesets that are charac-
%terizedbycompleteness,correctness,minimality,andnoneoverlappingtorepresenttheexplored
%domain.SuchrulesetsareconsideredastheoptimalresultsofLCSsandbeingtermedoptimalrule-
%set[O],whichareassumedtocontaininterpretablepatternsthatcanreflecttheunderlyingnature
%of the addressed problem (see Section 4). However, in practice, LCSs use EC techniques [38] as
%the primary search algorithms. The stochastic nature of EC makes LCSs prone to produce highly
%diverserules.ThisindicatesthatLCSscannotguaranteethecorrectnessandminimalityofthepro-
%duced ruleset. Furthermore, LCSs employ subsumption to prevent redundant specific rules to be
%introduced to the population [P]. Subsumption ensures that the evolved rules are unsubsumable
%ratherthanthattherulesarenotoverlapping.Thus,themajorityofLCSs-producedmodelsdonot
%containallthemember rulesof [O].

\section{Experiments on Binary Problems}
\label{sec:experiments_binary}
We performed a set of experiments to compare the size of the populations evolved by XCS, the number of niches identified by our approach, and the known optimal solution.

\subsection{Design of Experiments}
We followed the same experimental design and parameter settings used in the literature \cite{wilson:1995,wilson:1998,lanzi:1999:analysis,lanzi:2004:tec}. 
Each experiment consists of some learning problems that XCS must solve; each learning problem is followed by a test problem to evaluate XCS performance. During learning, XCS selects actions randomly, and the genetic algorithm is enabled; during testing, it performs the action with the highest prediction while the genetic algorithm is turned off. After learning is completed, we begin a condensation phase \citep{wilson:1995,10.1007/978-1-4471-0427-8_7,lanzi:2008:hyperellipsoidal} to compact the classifier population by turning off crossover and mutation during learning problems so that more general classifiers take over the population \cite{lanzi:2009:imbalanced} while no new classifier is created. In the experiments, we applied the basic XCS and therefore, to achieve optimal performance in challenging multi-step problems \cite{lanzi:1997:icga,lanzi:1999:analysis,lanzi:2004:tec}, we used\textit{ larger-than-usual populations} that were proved to produce optimal solutions \cite{lanzi:2004:tec}; all the other parameters were set as in the original experiments. The only exception was \W{14} that we could only solve using the parameter settings in \citep{lanzi:2004:tec}, coupling gradient descent with a biased exploration. We also gave XCS more opportunity for generalization with longer learning and condensation phases. Statistics are averages over 100 runs. 

\subsection{Boolean Functions}
In the first set of experiments, we applied XCS to two binary classification problems, Boolean Multiplexer (\MP{}) and Majority-on (\MAJ{})  (Section~\ref{sec:boolean_functions}) \cite{wilson:1995,10.1145/3468166}. The former is a well-known learning classifier systems test bed whose optimal solutions consists of few, highly general, and non-overlapping classifiers, accordingly, \CANP\ and \MANP\ should have very similar sizes. \MAJ{} poses a significant challenge to the vanilla XCS using ternary representation since its optimal solutions are represented by a number of highly overlapping (and highly specific) classifiers that is much larger than any other problems, as also predicted by the theoretical model of \cite{lanzi:2001:how}. Accordingly, \MAJ\  needs much larger populations than other problems of the same size while it can be comfortably solved with more advanced representations \cite{6595603}.

Table~\ref{tab:binary-combined}a compares the average population size, $|\CANP|$, and $|\MANP|$, before and after condensation, for \MP{} and \MAJ{} with different input sizes. For each problem, we report the population size $N$, the number of learning problems $n_{lp}$, the average population size ($|P|_{bc}$ and $|P|_{ac}|$), the number of active niches in the last population ($|\CAN|_{bc}$ and $|\CAN|_{ac}$), the mean number of niches in recorded populations ($|\MAN|_{bc}$ and $|\MAN|_{ac}$), and the size of the optimal population $|O|$ \cite{10.1145/3468166}. The reported statistics are averages over 100 runs with XCS settings that guaranteed optimal performance in all the runs.

Boolean multiplexer (\MP{}) have solutions consisting of highly general and non overlapping classifiers accordingly, as we previously anticipated, the size of \CANP\ and \MANP\ are basically identical at the end of the learning process, before condensation. After condensation, populations only contain the optimal classifiers so that, at the end, the population size, the number of currently active niches ($|\CAN|_{ac}$), and recorded niches ($|\MAN|_{ac}$) are identical and equal to $|O|$. 
In contrast, the optimal populations for Majority-On functions (\MAJ{}) consist of several highly overlapping, highly specific, classifiers so that XCS requires much larger populations to reach optimal performance. For example, while XCS needs a population of 400 classifiers to evolve the 16 optimal classifiers for \MP{6}, it needs a population of 2000 classifiers to evolve the 70 optimal classifiers for \MAJ{6}. Moreover, XCS needs 40000 classifiers to evolve the 924 optimal classifiers for \MAJ{10} while it only needs 5000 classifiers to solve the much larger \MP{37} optimally. 
For all the \MAJ{} problems, the number of currently active niches $|\CAN|$ is much lower than the mean number of niches recorded over the recent populations $|\MAN|$, both before and after condensation, since the classifiers in the population are highly overlapping. Simpler problems need smaller populations with fewer overlaps and thus have a lower $|\MAN|/|\CAN|$ ratio (1.5 and 1.23 for \MAJ{3} before and after condensation). Difficult problems needs larger population with much more overlaps and thus have a higher $|\MAN|/|\CAN|$  ratio (2.63 and 2.29 for \MAJ{10} before and after condensation). Note that in \MAJ, $|\CAN|$ is always lower than $|O|$ meaning at, on average, many of the classifiers in the population have been activated on overlapping niches (Figure~\ref{fig:overlapping_example}) and the number of distinct niche identifiers is much lower than the minimum number of niches $|O|$ needed to represent the optimal solution. However, although $|\MAN|$ is an estimate of the exact number of niche (Section~\ref{ssec:overlapping}), it seems to provide a reasonable evaluation of the number of active niches over the most recent iterations. 

\subsection{Multi-Step Problems}
We repeated the same comparison using well-known binary multi-step problems namely \W{1}, \W{14}, \W{2}, \MAZE{4}, \MAZE{5}, and \MAZE{6} \cite{wilson:1995,wilson:1998,lanzi:1997:icga,lanzi:1999:analysis}. These are grids containing impenetrable obstacles, goals, and empty positions; an agent must learn the shortest path to the goals from any empty position (Section~\ref{sec:grid_problems}). 
Table~\ref{tab:binary-combined}b reports, for each problem, the population size $N$, the number of learning problems $n_{lp}$, the average population size ($|P|_{bc}$ and $|P|_{ac}|$), the number of active niches in the last population ($|\CAN|_{bc}$ and $|\CAN|_{ac}$), the mean number of niches in recorded populations ($|\MAN|_{bc}$ and $|\MAN|_{ac}$), and the size of the optimal population $|O|$ which was recently computed in  \cite{lanzi:2025:optimal}. 
The reported statistics are averages over 100 runs. XCS parameters were set as in \cite{wilson:1995,wilson:1998} for \W{1} and \W{2}, as in \cite{lanzi:1997:icga} for \MAZE{4}, and as in \cite{lanzi:2004:tec} for \MAZE{5}, \MAZE{6}, and \W{14}. As previously mentioned, we did not employ any XCS extension but used larger-than-usual populations which \citet{lanzi:2004:tec} showed lead to optimal performance; only for \W{14} we used the exact setting of \cite{lanzi:2004:tec} combining gradient descent with biased exploration.

Overgeneral (and thus overlapping) classifiers are a known challenge for XCS in binary multi-step problems since the very beginning \cite{lanzi:1997:icga}. In fact, most of the work on multi-step problems have been focused on overcoming such a challenge \cite{lanzi:1999:analysis,lanzi:2004:tec,10.1007/3-540-45110-2_81,DBLP:conf/gecco/DrugowitschB05,10.1007/978-3-540-24855-2_90,6089622}. Table~\ref{tab:binary-combined}b shows that indeed during learning there is a considerable amount of overlapping classifiers in the populations: $|\CAN|_{bc}$ is always  lower than $|\MAN|_{bc}$ with a $|\MAN|/|\CAN|$ ratio around 1.5. 
Condensation gets rid of the overlapping classifiers and the values of $|\CAN|_{ac}$ and $|\MAN|_{ac}$ are basically identical and similar to $|O|$ suggesting once again that our approach provides a reasonable estimate of the number of active niches in the population. Interestingly, the values of $|\CAN|_{ac}$ for \MAZE{4}, \MAZE{5}, and \MAZE{6} are slightly lower than  the values of $|\MAN|_{ac}$, suggesting that the optimal solutions for these environments contains overlapping classifiers. 
We further analyzed the composition of the niches identified by our approach for the condensed solutions of these problems (Section~\ref{ssec:composition}) and discovered that indeed these contain overlapping classifiers. This is somehow surprising since it was assumed that these environment challenged XCS because they  allowed few generalizations \cite{lanzi:1997:icga,lanzi:1999:analysis}. However, not only they allow several generalizations (see \cite{lanzi:2025:optimal}) but we discovered that generalizations are also overlapping. Our approach also appears to be robust to the number of learning problems. As can be noted the evaluated numbers of niches ($|\CAN|$ and $|\MAN|$) remain basically the same after 5000 learning step, when XCS has already reached optimal performance and therefore is mainly working on discovering the most general classifiers \cite{lanzi:2004:tec}.

\section{Tracking Niche Evolution}
Our approach can identify the active niches in a populations using the $ats$ values which we can also use to compute (i) the niche composition (Section~\ref{ssec:composition}); (ii) the niche size, that is, the number of microclassifiers in the niche (Section~\ref{sec:xcs}); (iii) the niche average fitness; and (iv) to track the evolution of niches over time. As an example, we applied XCS to \MP{20} with the same settings used in Table~\ref{tab:binary-combined}a for 100000 learning problems; we did not include the condensation phase since it is less interesting in terms of niche evolution.

\begin{figure}
	\subfloat[]{
	\centering
	\includegraphics[width=.33\textwidth]{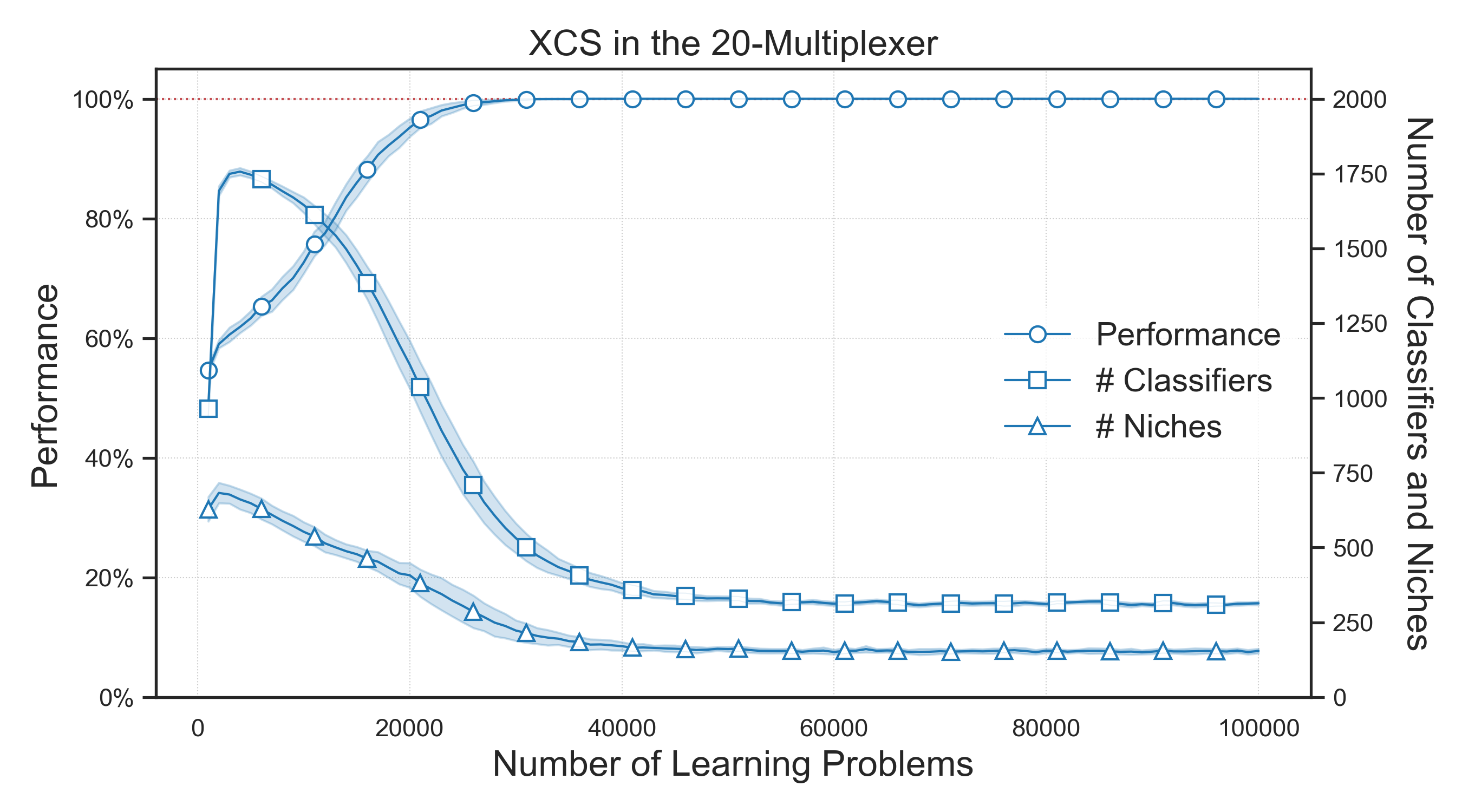}
	\label{fig:mp20_performance}
}

\subfloat[]{
	\centering
	\includegraphics[width=.33\textwidth]{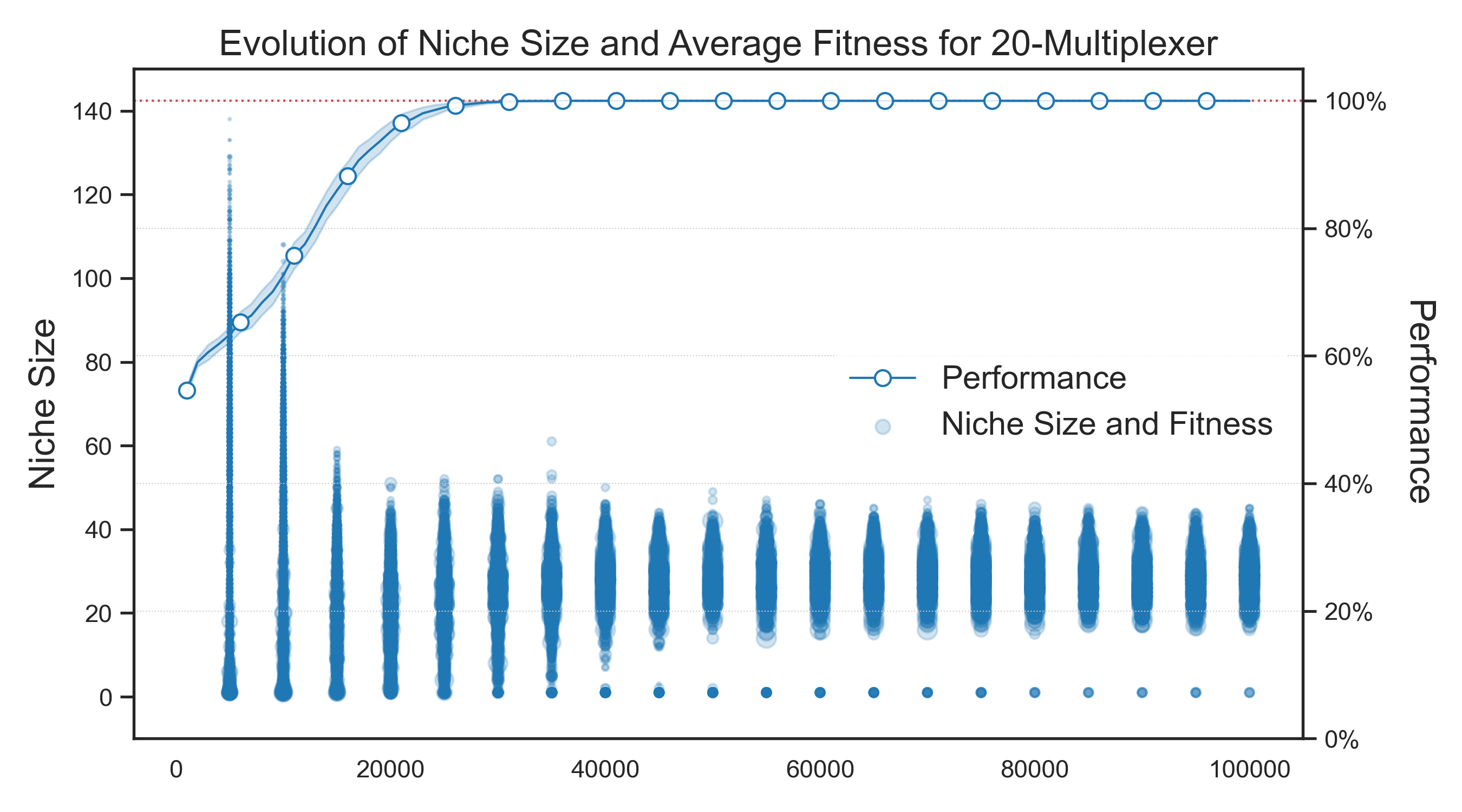}
	\label{fig:mp20_niche_size}
}
	
	\caption{XCS applied to the 20-multiplexer: (a) performance, number of classifiers, and currently active niches (\CAN), curves are averages over 20 runs); (b) evolution of niche size using the data collected of all the 20 runs.}
	\label{fig:mp20_niches}
\end{figure}

Figure~\ref{fig:mp20_performance} reports the performance, the number of classifiers in the population, and the number of currently active niches $|\CANP|$, which in the case of \MP{20} is basically identical to $\MANP|$ (Section \ref{ssec:overlapping}); curves are averages over 20 runs; the colored area show the confidence interval ($\mu\pm\sigma$). The results are coherent with what widely known in the literature \cite{wilson:1998,lanzi:2004:tec}. XCS reaches optimal performance around 40000 learning problems. Initially, the population contains more specific classifiers and the number of classifiers grows (Figure~\ref{fig:mp20_performance}); as learning continues, the genetic algorithm favors the evolution of accurate maximally general classifiers and the population shrinks to a much smaller number of more general classifiers, also thanks to the subsumption deletion operator \cite{wilson:1998}.
At the end, the population contains the 64 classifiers that represent \MP{20} optimal solution \cite{wilson:1998}
and other classifiers that are continually generated through recombination and mutation, which would be eliminated by applying condensation (Table~\ref{tab:binary-combined}).
This behavior is reflected in the plot of the number of niches that our approach identifies (triangle markers in Figure~\ref{fig:mp20_performance}). 
The number of active niches $|\CANP|$ initially increases since specific classifiers tend to cover small areas of the problem space. 
Then, the evolution focuses on more general classifiers and the number of niches decrease and becomes stable around an average of $154.30\pm7.67$ niches (around twice the minimum number of niches need to solve the problem). 

%%
%We can use the information collected from the action set time stamps to analyze the evolution of size, average fitness, and amount of overlap of active niches during learning (Section~\ref{sec:niche_evolution}). 
%
Figure~\ref{fig:mp20_niche_size} reports the size and average fitness of the active niches for the intermediate populations collected every 5000 learning problems, \textit{over all the 20 runs}. Each vertical column in Figure~\ref{fig:mp20_niche_size} shows the distribution of niche size\textit{ collected over all the 20 intermediate populations}. A dot represents an active niche; its position on the $x$ axis is the learning problem when the corresponding population was saved; its position on the $y$ axis is the niche size computed as the number of microclassifiers in the niche (Section~\ref{sec:xcs} and \ref{ssec:composition}); the dot size represents the average fitness of the classifiers in the same niche: the larger the dot, the higher the average fitness. 
Figure~\ref{fig:mp20_niche_size} shows that, initially, there are many active niches; some of them contain many low fitness (thus over general) classifiers, represented by the tiny dots at the top of the first two segments containing more than 100 classifiers; other ones contain fewer high fitness (thus more specific) classifiers, represented by the large dots at the bottom of the first two vertical segments.  As learning progress, XCS favors the evolution of accurate maximally general classifiers, accordingly, overgeneral classifiers are deleted and the size of active niches decreases while specific classifiers are deleted as they appear in fewer niches and thus they are reproduced less. When XCS reaches optimal performance (around 40000 learning problems), the evolution is focused on fewer (Figure~\ref{fig:mp20_performance}), smaller, high fitness niches, as shown by the narrower distribution of large dots. Note that, there are still larger lower fitness niches at the top of the distributions and smaller higher fitness niches at the bottom; XCS is still learning and the genetic algorithm can still generate over general and over specific classifiers.

%%\begin{figure}[t]
%%	\subfloat{\includegraphics[width=\columnwidth]{figures/xcs/fig_mp20_performance_population_niches.png}}		
%%	
%%	\subfloat{\includegraphics[width=\columnwidth]{figures/xcs/fig_mp20_percentages_overlaps_area.png}}		
%%	
%%	\subfloat{\includegraphics[width=\columnwidth]{figures/xcs/fig-mp20-niche-size-evolution-with-performance-no-experiments-0020-selection-005000.png}}		
%%	
%%	%
%%	%\includegraphics[width=.9\columnwidth]{figures/fig_mp20_percentages_overlaps_area.png}
%%	%
%%	%\includegraphics[width=.9\columnwidth]{figures/fig-mp20-niche-size-evolution.png}
%%	%
%%	\caption{XCS applied to the 20-multiplexer: performance, number of classifiers and niches. Curves are averages over 20 runs.}
%%	\label{fig:xcs_mp20}
%%\end{figure}
%
%\begin{figure}
%	\subfigure{\includegraphics[width=\columnwidth]{figures/xcs/fig_mp20_performance_population_niches.png}}
%	\caption{XCS applied to the 20-multiplexer: performance, number of classifiers and niches. Curves are averages over 20 runs.}
%	\label{fig:xcs_mp20_ppn}
%\end{figure}
%	
%\begin{figure}
%	\subfigure[ ]{\includegraphics[width=\columnwidth]{figures/xcs/fig-mp20-niche-size-evolution-with-performance-no-experiments-0020-selection-005000.png}}		
%
%	\subfigure[ ]{\includegraphics[width=\columnwidth]{figures/xcs/fig_mp20_percentages_overlaps_area.png}}		
%	
%	\caption{XCS applied to the 20-multiplexer: (a) percentage of niche overlaps; (b) evolution of niche size. Plots report the data of 20 runs.}
%	\label{fig:xcs_mp20_niches}
%\end{figure}

\section{Conclusions}
We presented an approach to identify the currently active niches in evolving classifier populations that solely relies on XCS occurrence-based niched genetic algorithm and, therefore, can be applied to any existing XCS model with minimal modifications. We applied our approach to binary single-step and multi-step problems to evaluate the number of niches in populations during learning and after the population has been minimized through condensation. We compared the number of active niches our method estimates with the known optimal number of niches (i.e., the number of classifiers needed to represent the optimal solution \cite{lanzi:2001:how}). We show that our approach can provide a reliable estimate in problems whose optimal solutions only comprise non-overlapping classifiers and in challenging problems involving optimal populations of massively overlapping classifiers. Our approach can also highlight the presence of overlapping classifiers without performing any matching operation but just comparing the number of niches evaluated during the recently tracked populations. Our approach can also be used to compute the composition of active niches and track their evolution using useful statistics such as the niche size, average fitness, etc. As for future research directions, there is still much unexplored potential since, for instance, this analysis could be extended to more advanced representations (e.g., code fragments \cite{6595603}). We also did not fully exploit the knowledge of niche composition, which could be used to improve existing compaction algorithms and lead to more explainable classifier systems models. 

%All the code and configuration files to replicate the results are available at \cite{duplication_github}; actual data generated for the paper is available on request but the amount of hard disk space required is substantial. 

\bibliographystyle{ACM-Reference-Format}
\bibliography{./bibliography/gecco2024-niches,./bibliography/lanzi,./bibliography/gecco2025-lanzi-anonymous}   % name your BibTeX data base
\begin{table*}
	%		\caption{Reports the STANDARD DEVIATION OF THE SIGNATURE MEANS}
	\begin{tabular}{|c|c|c||c|c|c||c|c|c||c|}\hline
		{\bf Problem} & $N$ & $n_{lp}$
		
		%%% PRE 
		& $|P|_{bc}$ & $|\CAN|_{bc}$ & $|\MAN|_{bc}$ 
		
		%%% POST
		& $|P|_{ac}$ & $|\CAN|_{ac}$ & $|\MAN|_{ac}$ 
		
		& $|O|$ \\ \hline
		
		\MP{6} &        400  &      10000  & $  27.7 \pm    3.5$  & $  18.9 \pm    1.5$  & $  20.0 \pm    2.1$  & $  16.0 \pm    0.0$  & $  16.0 \pm    0.0$  & $  16.0 \pm    0.0$  &     16 \\
		\MP{11} &       1000  &      20000  & $  86.0 \pm    7.2$  & $  47.4 \pm    3.4$  & $  47.4 \pm    3.4$  & $  32.0 \pm    0.0$  & $  32.0 \pm    0.0$  & $  32.0 \pm    0.0$  &     32 \\
		\MP{20} &       2000  &     200000  & $ 270.1 \pm   15.4$  & $ 129.6 \pm    7.7$  & $ 129.6 \pm    7.7$  & $  64.0 \pm    0.0$  & $  64.0 \pm    0.0$  & $  64.0 \pm    0.0$  &     64 \\
		\MP{37} &       5000  &    1000000  & $1472.8 \pm   31.2$  & $ 576.1 \pm   17.6$  & $ 576.1 \pm   17.6$  & $ 128.0 \pm    0.0$  & $ 128.0 \pm    0.0$  & $ 128.0 \pm    0.0$  &    128 \\
		\MAJ{3} &        500  &      10000  & $  23.7 \pm    2.6$  & $  13.8 \pm    1.4$  & $  21.5 \pm    2.2$  & $  16.2 \pm    2.7$  & $  12.9 \pm    1.1$  & $  15.9 \pm    2.5$  &     12 \\
		\MAJ{4} &       1000  &      10000  & $  54.5 \pm    6.0$  & $  24.2 \pm    2.4$  & $  45.0 \pm    4.5$  & $  33.9 \pm    4.2$  & $  22.5 \pm    2.2$  & $  33.1 \pm    4.0$  &     20 \\
		\MAJ{5} &       2000  &      10000  & $ 115.8 \pm   11.9$  & $  43.7 \pm    4.6$  & $  88.0 \pm    8.8$  & $  70.6 \pm    8.2$  & $  40.3 \pm    4.0$  & $  68.4 \pm    7.9$  &     40 \\
		\MAJ{6} &       2000  &      20000  & $ 232.8 \pm   16.5$  & $  76.3 \pm    5.9$  & $ 164.6 \pm   11.6$  & $  92.3 \pm    5.2$  & $  63.9 \pm    4.6$  & $  90.4 \pm    5.1$  &     70 \\
		\MAJ{7} &       4000  &      40000  & $ 553.5 \pm   25.4$  & $ 151.3 \pm    8.5$  & $ 357.5 \pm   16.7$  & $ 176.7 \pm    6.8$  & $ 121.0 \pm    5.9$  & $ 173.2 \pm    6.6$  &    140 \\
		\MAJ{8} &       8000  &      80000  & $1098.1 \pm   47.9$  & $ 263.8 \pm   11.5$  & $ 656.4 \pm   29.4$  & $ 318.5 \pm    9.2$  & $ 204.5 \pm    8.6$  & $ 310.6 \pm    9.1$  &    252 \\
		\MAJ{9} &      16000  &      80000  & $1803.8 \pm   76.9$  & $ 369.4 \pm   20.4$  & $ 930.5 \pm   46.5$  & $ 679.1 \pm   18.0$  & $ 292.6 \pm   17.6$  & $ 622.4 \pm   22.7$  &    504 \\
		\MAJ{10} &      40000  &     200000  & $3784.8 \pm  101.9$  & $ 658.4 \pm   27.2$  & $1730.0 \pm   58.0$  & $1285.4 \pm   23.9$  & $ 508.6 \pm   20.2$  & $1166.8 \pm   39.4$  &    924 \\\hline

	\end{tabular}
	
	\vskip.25cm		
	\centerline{(a)}
	
	\vskip.5cm

	\begin{tabular}{|c|c|c||c|c|c||c|c|c||c|}\hline
		{\bf Problem} & $N$ & $n_{lp}$
		
		%%% PRE 
		& $|P|_{bc}$ & $|\CAN|_{bc}$ & $|\MAN|_{bc}$ 
		
		%%% POST
		& $|P|_{ac}$ & $|\CAN|_{ac}$ & $|\MAN|_{ac}$
		
		& $|O|$ \\ \hline
		
		\W{1} &        800  &       5000  & $ 114.2 \pm   10.5$  & $  46.4 \pm    3.4$  & $  67.2 \pm    4.8$  & $  32.0 \pm    1.0$  & $  31.8 \pm    1.0$  & $  32.0 \pm    1.0$  &     31 \\
		\W{1} &        800  &      10000  & $ 112.4 \pm   12.9$  & $  46.3 \pm    3.2$  & $  67.5 \pm    5.7$  & $  31.6 \pm    0.6$  & $  31.5 \pm    0.6$  & $  31.6 \pm    0.6$  &     31 \\
		\W{1} &        800  &      20000  & $ 110.8 \pm   10.6$  & $  46.0 \pm    3.1$  & $  66.6 \pm    4.5$  & $  31.6 \pm    0.6$  & $  31.6 \pm    0.6$  & $  31.6 \pm    0.6$  &     31 \\\hline
%		\W{1} &       1600  &       5000  & $ 196.2 \pm   16.2$  & $  61.6 \pm    4.0$  & $  98.9 \pm    6.4$  & $  33.1 \pm    1.3$  & $  31.5 \pm    0.7$  & $  31.7 \pm    0.8$  &     31 \\
%		\W{1} &       1600  &      10000  & $ 196.3 \pm   15.4$  & $  61.5 \pm    3.5$  & $  99.2 \pm    6.2$  & $  31.5 \pm    0.7$  & $  31.3 \pm    0.6$  & $  31.4 \pm    0.6$  &     31 \\
%		\W{1} &       1600  &      20000  & $ 191.6 \pm   15.0$  & $  61.5 \pm    3.8$  & $  97.6 \pm    5.6$  & $  31.2 \pm    0.4$  & $  31.2 \pm    0.4$  & $  31.2 \pm    0.4$  &     31 \\\hline
		\W{2} &        800  &       5000  & $ 130.6 \pm   13.4$  & $  48.5 \pm    3.3$  & $  72.2 \pm    5.4$  & $  31.8 \pm    0.9$  & $  31.7 \pm    0.9$  & $  31.8 \pm    0.9$  &     31 \\
		\W{2} &        800  &      10000  & $ 124.7 \pm   12.8$  & $  48.2 \pm    3.4$  & $  72.3 \pm    5.7$  & $  31.6 \pm    0.8$  & $  31.5 \pm    0.8$  & $  31.6 \pm    0.8$  &     31 \\
		\W{2} &        800  &      20000  & $ 121.2 \pm   10.1$  & $  48.2 \pm    3.0$  & $  71.4 \pm    5.0$  & $  31.2 \pm    0.5$  & $  31.1 \pm    0.5$  & $  31.2 \pm    0.5$  &     31 \\\hline
%		\W{2} &       1600  &       5000  & $ 228.4 \pm   20.5$  & $  68.0 \pm    4.5$  & $ 112.7 \pm    7.7$  & $  33.4 \pm    1.7$  & $  31.6 \pm    0.8$  & $  31.8 \pm    0.8$  &     31 \\
%		\W{2} &       1600  &      10000  & $ 219.4 \pm   17.5$  & $  67.4 \pm    4.0$  & $ 111.7 \pm    7.3$  & $  31.6 \pm    0.7$  & $  31.4 \pm    0.6$  & $  31.5 \pm    0.6$  &     31 \\
%		\W{2} &       1600  &      20000  & $ 220.9 \pm   18.8$  & $  67.5 \pm    4.7$  & $ 111.3 \pm    6.8$  & $  31.2 \pm    0.4$  & $  31.2 \pm    0.4$  & $  31.2 \pm    0.4$  &     31 \\\hline
		\MAZE{4} &       2000  &       5000  & $ 815.8 \pm   21.7$  & $ 142.7 \pm    4.8$  & $ 231.0 \pm    8.2$  & $ 100.0 \pm    0.8$  & $  97.8 \pm    1.6$  & $  99.8 \pm    0.8$  &     99 \\
		\MAZE{4} &       2000  &      10000  & $ 816.8 \pm   19.8$  & $ 143.4 \pm    4.5$  & $ 233.2 \pm    8.8$  & $  99.5 \pm    0.7$  & $  97.4 \pm    1.3$  & $  99.4 \pm    0.7$  &     99 \\
		\MAZE{4} &       2000  &      20000  & $ 811.3 \pm   19.2$  & $ 142.9 \pm    4.6$  & $ 231.5 \pm    8.7$  & $  99.3 \pm    0.7$  & $  97.5 \pm    1.4$  & $  99.1 \pm    0.7$  &     99 \\
		\MAZE{4} &       2000  &      40000  & $ 815.8 \pm   18.5$  & $ 142.5 \pm    4.8$  & $ 231.1 \pm    9.2$  & $  99.0 \pm    0.6$  & $  97.4 \pm    1.5$  & $  98.9 \pm    0.7$  &     99 \\\hline
%		\MAZE{5} &       3500  &       5000  & $1453.5 \pm   28.2$  & $ 235.1 \pm    5.4$  & $ 366.3 \pm   12.8$  & $ 179.0 \pm    1.3$  & $ 176.5 \pm    2.0$  & $ 178.9 \pm    1.3$  &    176 \\
%		\MAZE{5} &       3500  &      10000  & $1457.3 \pm   28.5$  & $ 235.2 \pm    5.0$  & $ 364.8 \pm   13.6$  & $ 178.7 \pm    1.2$  & $ 176.2 \pm    1.8$  & $ 178.5 \pm    1.2$  &    176 \\
%		\MAZE{5} &       3500  &      20000  & $1461.7 \pm   28.6$  & $ 235.2 \pm    5.0$  & $ 365.9 \pm   12.3$  & $ 178.4 \pm    1.2$  & $ 176.4 \pm    1.8$  & $ 178.3 \pm    1.2$  &    176 \\
%		\MAZE{5} &       3500  &      40000  & $1455.7 \pm   28.6$  & $ 235.9 \pm    6.1$  & $ 368.0 \pm   11.9$  & $ 178.1 \pm    1.6$  & $ 176.0 \pm    2.0$  & $ 178.0 \pm    1.6$  &    176 \\\hline
		\MAZE{5} &       4000  &       5000  & $1763.3 \pm   33.7$  & $ 243.9 \pm    4.8$  & $ 389.9 \pm   15.4$  & $ 178.5 \pm    1.3$  & $ 175.3 \pm    2.0$  & $ 178.4 \pm    1.3$  &    176 \\
		\MAZE{5} &       4000  &      10000  & $1763.5 \pm   28.1$  & $ 244.4 \pm    5.1$  & $ 394.0 \pm   13.1$  & $ 178.3 \pm    1.2$  & $ 174.7 \pm    1.9$  & $ 178.2 \pm    1.2$  &    176 \\
		\MAZE{5} &       4000  &      20000  & $1764.4 \pm   30.9$  & $ 244.3 \pm    4.7$  & $ 391.2 \pm   12.7$  & $ 178.1 \pm    1.0$  & $ 175.3 \pm    1.9$  & $ 178.0 \pm    1.0$  &    176 \\
		\MAZE{5} &       4000  &      40000  & $1755.8 \pm   27.8$  & $ 242.8 \pm    5.1$  & $ 390.2 \pm   14.0$  & $ 178.1 \pm    1.0$  & $ 175.3 \pm    1.9$  & $ 178.0 \pm    1.0$  &    176 \\\hline
		\MAZE{6} &       4000  &       5000  & $1793.4 \pm   40.0$  & $ 236.1 \pm    5.3$  & $ 388.6 \pm   18.0$  & $ 165.8 \pm    0.6$  & $ 161.7 \pm    1.6$  & $ 165.6 \pm    0.6$  &    165 \\
		\MAZE{6} &       4000  &      10000  & $1794.5 \pm   39.9$  & $ 235.8 \pm    5.2$  & $ 388.6 \pm   18.1$  & $ 165.5 \pm    0.7$  & $ 161.8 \pm    1.6$  & $ 165.3 \pm    0.6$  &    165 \\
		\MAZE{6} &       4000  &      20000  & $1789.8 \pm   43.3$  & $ 237.0 \pm    5.7$  & $ 388.5 \pm   19.4$  & $ 165.1 \pm    0.5$  & $ 161.9 \pm    1.6$  & $ 164.9 \pm    0.5$  &    165 \\
		\MAZE{6} &       4000  &      40000  & $1792.3 \pm   37.5$  & $ 237.2 \pm    4.7$  & $ 390.2 \pm   18.3$  & $ 165.1 \pm    0.6$  & $ 162.1 \pm    1.6$  & $ 165.0 \pm    0.6$  &    165 \\\hline
		\W{14} &       4000  &       5000  & $ 926.6 \pm   86.5$  & $ 139.9 \pm    2.7$  & $ 157.2 \pm    4.8$  & $ 307.9 \pm   40.8$  & $ 138.4 \pm    3.2$  & $ 140.9 \pm    2.0$  &    137 \\
		\W{14} &       4000  &      10000  & $1248.5 \pm   85.9$  & $ 142.8 \pm    1.0$  & $ 154.4 \pm    3.0$  & $ 255.9 \pm   19.4$  & $ 140.2 \pm    1.5$  & $ 140.6 \pm    1.4$  &    137 \\
		\W{14} &       4000  &      20000  & $1402.2 \pm   58.3$  & $ 143.2 \pm    0.8$  & $ 153.4 \pm    2.9$  & $ 198.9 \pm    7.4$  & $ 139.9 \pm    1.3$  & $ 140.1 \pm    1.3$  &    137 \\
		\W{14} &       4000  &      40000  & $1422.2 \pm   57.1$  & $ 143.3 \pm    0.8$  & $ 152.4 \pm    2.6$  & $ 157.7 \pm    4.1$  & $ 139.8 \pm    1.3$  & $ 139.9 \pm    1.3$  &    137 \\\hline		
	\end{tabular}	
	
\vskip.25cm		
	\centerline{(b)}

		\vskip.5cm		
	\caption{Population size evolved by XCS compared to the number of currently active niches \CANP\ and the mean number of active niches in recorded populations \MANP, before and after condensation: 
		(a) binary single-step problems;
		(b) grid environments;
		statistics are averages over 100 runs reported using mean and standard deviation ($\mu\pm\sigma$).  
		For each problem we report 
		(i) the population size $N$;
		(ii) the number of learning problems $n_{lp}$;
		(iii) the population size before and after condensation ($|P|_{bc}$, $|P|_{ac}$);
		(iv) the number of the active niches before  and after condensation ($|\CAN|_{bc}$, $|\CAN|_{ac}$);
		(v) the mean number of the active niches in recorded populations before and after condensation ($|\MAN|_{bc}$, $|\MAN|_{ac}$) ;
		(vi) the number of the optimal (accurate, maximally general) populations for the same problem ($|O|$) that is also the minimum number of niches needed to evolve the optimal solution \cite{lanzi:2001:how}.
	}	
	\label{tab:binary-combined}
\end{table*}

\appendix
\section{Boolean Functions}
\label{sec:boolean_functions}
Boolean functions are defined over binary strings of size $n$ and return the truth value associated to it. They can be modeled as reinforcement learning problems \citep{wilson:1995}; at each time step, the system receives a binary string and outputs a binary action; if the action corresponds to the function output, the system receives a reward of 1000, otherwise 0. 

Boolean multiplexer (\MP{}) are defined over binary strings of length $n$ with $n=k+2^k$; the first $k$ bits represent an address to the remaining $2^k$ bits; the function returns the value of the indexed bit. For example, the multiplexer of size 6 (\MP{6}) applied to $110001$ returns $1$ since the first 2 bits ($11$) point to the fourth position of the following four bits ($0001$); similarly, \MP{6}($000111$) returns $0$. 
%More formally, the 6-multiplexer can be defined as, 
%%
%\begin{equation}	
%	\protect{mp}_6(x_0, x_1, y_0, \dots, y_3) = 
%	\overline{x_0}~\overline{x_1}~ y_0 + \overline{x_0}~x_1~y_1  + x_0~\overline{x_1}~y_2 + x_0~x_1~y_3 \label{eq:mp6}
%\end{equation}

%In carry problems (\CAR{}), the input string represents
%two binary numbers of the same length ($n/2$)
%and output the carry bit generated by their sum. 
%For example, the output of \CAR{6} for 100101 is 1 while for 100001 is 0. 
%Carry problems are challenging for XCS since their solution involve 
%overlapping classifiers and evolutionary niches 
%which often maintain over-general classifier in population \citep{DBLP:conf/gecco/LiuB022}.

Majority-on functions (\MAJ{}) output the most frequent bit of the input sequence that is, 
they output 1 when the majority of the input bits are set to one, 0 otherwise. 
\MAJ{} problems are challenging since the optimal solutions are represented by a number of classifiers that 
is much larger than any other problems. For example, the optimal solution for \MAJ{11} consists of 
1848 classifiers while the optimal solution for \MP{70} only needs 256 classifiers \citep{DBLP:conf/gecco/LiuB022,10.1145/3468166}.

\section{Grid Problems}
\label{sec:grid_problems}
\begin{figure*}[h]
	\subfloat[\W{1}]{
		\centering
		\includegraphics[height=1.3in]{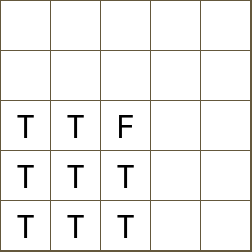}
		\label{fig:environment_woods1}
	}
	\subfloat[\W{14}]{
	\centering
	\includegraphics[height=1.3in]{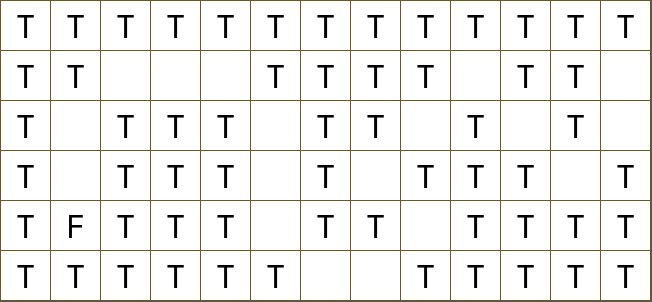}
	\label{fig:environment_woods14}
}
	\subfloat[\MAZE{4}]{
		\centering
		\includegraphics[height=1.3in]{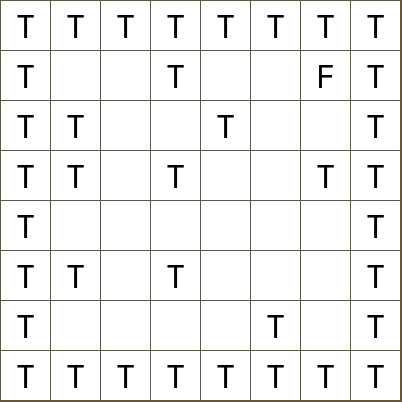}
		\label{fig:environment_maze4}
	}

\vskip.5cm

	\subfloat[\W{2}]{
		\centering
		\includegraphics[height=1.3in]{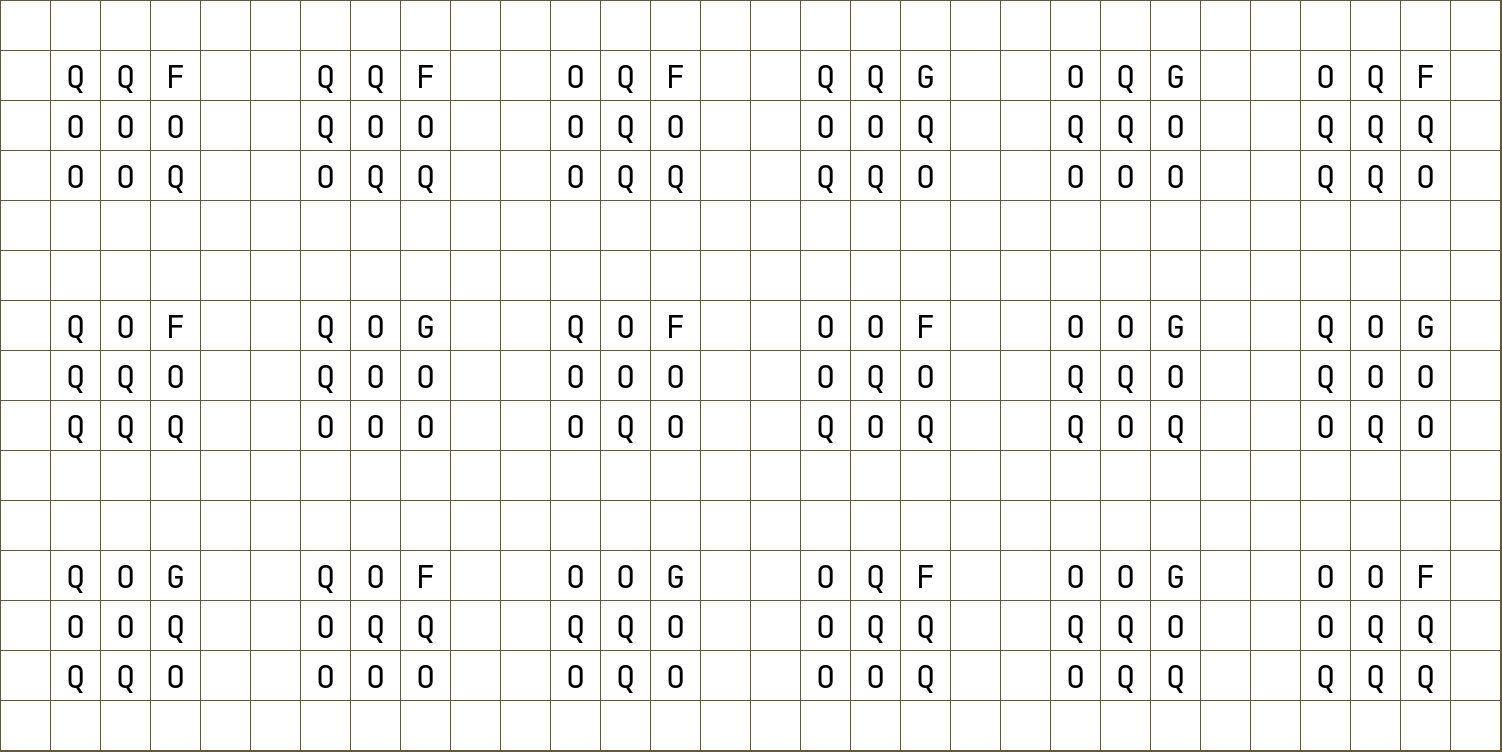}
		\label{fig:environment_woods2}
	}
	\subfloat[\MAZE{5}]{
		\centering
		\includegraphics[height=1.3in]{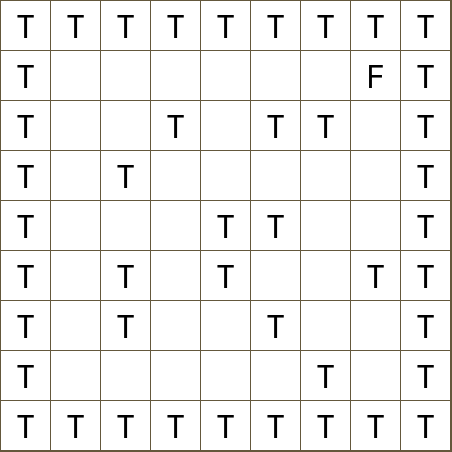}
		\label{fig:environment_maze5}
	}
	\subfloat[\MAZE{6}]{
		\centering
		\includegraphics[height=1.3in]{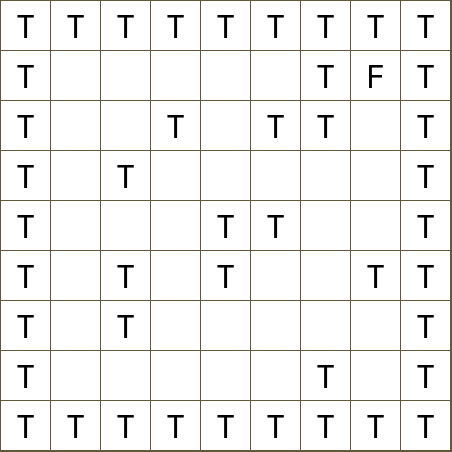}
		\label{fig:environment_maze6}
	}

%	
%	
%	\begin{tabular}{ccc}
%		\multicolumn{3}{c}{\parbox{\linewidth}{\centerline{\includegraphics[height=1.8in]{figures/environments/woods2}}}}\\
%		& & \\
%		\multicolumn{3}{c}{(a) \W{2}} \\
%		& & \\
%		\parbox{.3\linewidth}{\centerline{\includegraphics[height=1.3in]{figures/environments/maze4}}} &
%		\parbox{.3\textwidth}{\centerline{\includegraphics[height=1.3in]{figures/environments/maze5}}} & 
%		\parbox{.3\textwidth}{\centerline{\includegraphics[height=1.3in]{figures/environments/maze6}}} \\
%		& & \\
%		(b) \MAZE{4} & (c) \MAZE{5} & (d) \MAZE{6} \\
%		& & \\		
%		\multicolumn{3}{c}{\centerline{\includegraphics[height=1.0in]{figures/environments/woods14}}} \\
%		& & \\		
%		\multicolumn{3}{c}{(e) \W{14}} \\
%		& & \\
%	\end{tabular}	
	\caption{Grid environments that have been used for evaluating XCS performance in multistep problems:
		(a) \W{1}, (b) \W{14}, (c) \MAZE{4}, (d) \W{2}, (e) \MAZE{5}, and (f) \MAZE{6} \citep{wilson:1995,lanzi:1997:icga,wilson:1998,lanzi:1999:analysis}.}
	\label{fig:grids}
\end{figure*}

These are grids with impenetrable obstacles (``\texttt{T}''), goals (``\texttt{F}''), and empty positions (``\texttt{ }'').
The system can occupy empty positions and move to any adjacent empty position or goal position.
It has eight binary sensors, one for each adjacent position, 
that encode obstacles as \texttt{10}, goals as \texttt{11}, and empty positions as \texttt{00}.
Thus, its sensory input is a string of 16 bits (2 bits $\times$ 8 positions).	
The agent can perform eight actions ($a_0, \dots, a_7$), 
one for each possible adjacent position, encoded as three-bit strings ($a_0$ as \texttt{000}, \dots, $a_7$ as \texttt{111});
$a_0$ corresponds to moving north (up), and the next ones identify the remaining directions following a clockwise order.
The agent receives a 1000 reward when it reaches the goal, zero otherwise.
%{fig:grids}
\W{1} (\ref{fig:environment_woods1}) \citep{wilson:1995} is a toroidal grid with 16 empty positions and one goal position; \W{14} (\ref{fig:environment_woods14}) is a long corridor created to challenge long action chains; 
\MAZE{4} , \MAZE{5}, and \MAZE{6}  (\ref{fig:environment_maze4}-\ref{fig:environment_maze6}) are mazes created to challenge generalization; all these environments share the same sensor model using 16 bits to encode the eight surrounding positions and 3 bits for the actions. \W{2} (\ref{fig:environment_woods2}) \cite{wilson:1998} extends such sensor model with two types of goals (``F'' and ``G'') and two types of obstacles (``Q'' and ``O''); each sensors is encoded using three bits, therefore the sensory input is a string of 24 bits.  The environment is structurally similar to \W{1} but has a larger input space that, as \cite{wilson:1998} showed, allows for more generalizations.
The images of the grid environments are available in Figure~\ref{fig:environment_woods1}-\ref{fig:environment_maze6}.

%\begin{figure*}[h]
%	\begin{tabular}{ccc}
%		\multicolumn{3}{c}{\parbox{\linewidth}{\centerline{\includegraphics[height=1.8in]{figures/environments/woods2}}}}\\
%		& & \\
%		\multicolumn{3}{c}{(a) \W{2}} \\
%		& & \\
%		\parbox{.3\linewidth}{\centerline{\includegraphics[height=1.3in]{figures/environments/maze4}}} &
%		\parbox{.3\textwidth}{\centerline{\includegraphics[height=1.3in]{figures/environments/maze5}}} & 
%		\parbox{.3\textwidth}{\centerline{\includegraphics[height=1.3in]{figures/environments/maze6}}} \\
%		& & \\
%		(b) \MAZE{4} & (c) \MAZE{5} & (d) \MAZE{6} \\
%		& & \\		
%		\multicolumn{3}{c}{\centerline{\includegraphics[height=1.0in]{figures/environments/woods14}}} \\
%		& & \\		
%		\multicolumn{3}{c}{(e) \W{14}} \\
%		& & \\
%	\end{tabular}	
%	\caption{Grid environments that have been used for evaluating XCS performance in multistep problems:
%		(a) \W{2}, (b) \MAZE{4}, (c) \MAZE{5}, (d) \MAZE{6}, and (e) \W{14} \citep{wilson:1995,wilson:1998,lanzi:1997:icga,lanzi:1999:analysis}.}
%	\label{fig:grids}
%\end{figure*}
\clearpage

\end{document}